\pdfoutput=1

\documentclass[11pt]{article}

\usepackage[final]{acl}

\usepackage{times}
\usepackage{latexsym}
\usepackage{graphicx}    
\usepackage{caption}     
\usepackage{subcaption}  
\usepackage[T1]{fontenc}

\usepackage[utf8]{inputenc}

\usepackage{microtype}

\usepackage{inconsolata}

\usepackage{graphicx}
\usepackage{todonotes}
\usepackage{color}
\definecolor{purple}{rgb}{1,0,1}
\newcount\Comments  
\newcommand{\kibitz}[2]{\ifnum\Comments=1{\color{#1}{#2}}\fi}

\newcommand{\OUT}[1]{}

\title{Towards conversational assistants for health applications: using ChatGPT to generate conversations about  heart failure}
\author{
Anuja Tayal\textsuperscript{1} \quad
Devika Salunke \textsuperscript{2}\quad
Barbara Di Eugenio\textsuperscript{1} \quad 
Paula G Allen-Meares \textsuperscript{3} \\
\textbf{
Eulalia P Abril \textsuperscript{4} \quad
Olga Garcia-Bedoya\textsuperscript{3} \quad
Carolyn A Dickens\textsuperscript{3} \quad
Andrew D. Boyd\textsuperscript{2}} \\
\textsuperscript{1}Department of Computer Science 
\textsuperscript{2}Department of Biomedical and Health Information Sciences \\
\textsuperscript{3}Department of Medicine
\textsuperscript{4}Department of Communications \\
\texttt{\{atayal4,dsalun2,bdieugen,pameares,eulalia,ogarciab,cdickens,boyda\}@uic.edu} \\
}
\begin{document}
\maketitle
\begin{abstract}

We explore the potential of ChatGPT (3.5-turbo and 4) to generate conversations focused on self-care strategies for African-American heart failure patients—a domain with limited specialized datasets. To simulate patient-health educator dialogues, we employed four prompting strategies: domain, African American Vernacular English (AAVE), Social Determinants of Health (SDOH), and SDOH-informed reasoning. Conversations were generated across key self-care domains—food, exercise, and fluid intake—with varying turn lengths (5, 10, 15)  and incorporated patient-specific SDOH attributes such as age, gender, neighborhood, and socioeconomic status. Our findings show that effective prompt design is essential. While incorporating SDOH and reasoning improves dialogue quality, ChatGPT still lacks the empathy and engagement needed for meaningful healthcare communication.

\end{abstract}
\section{Introduction}
Heart failure (HF), or congestive heart failure, occurs when the heart cannot pump enough blood to meet the body’s needs. Effective self-care—such as managing salt intake, staying hydrated, exercising, adhering to medications, and attending regular check-ups—is critical for managing the condition \cite{hf-history}. However, African Americans (AA) in the U.S. face disproportionately worse outcomes due to genetic factors, limited healthcare access, socioeconomic challenges, and lower health literacy \cite{hfaa}. Existing self-care materials often cater to a white, educated population, lacking cultural relevance for minority communities \cite{hf-challenges}. This gap contributes to poor adherence and worsened outcomes. Personalized education can improve self-care understanding and reduce readmissions \cite{self-care-design}.

This study is part of a broader project aimed at developing a culturally sensitive conversational agent to support AA heart failure patients in asking self-care related questions. A significant challenge is the lack of real-world patient-centered conversational data from underrepresented communities. To address this, we recruited three patient educators (PEs) to provide heart failure education to 18 AA and 2 Hispanic/Latino (H/L) patients (Reference withheld).  Initial analysis revealed that educators dominated the conversations, with patients contributing less. The key topics discussed during these sessions included exercise, fluid intake, symptom management, sleep, weight management, familial aspects, and salt intake. 

Unlike the recorded interactions, we wanted to generate conversational datasets that are initiated by patients and, more importantly, personalized based on the Social Determinants of Health (SDOH) features. With the advancement of Large Language Models (LLMs), we thought to examine different prompting strategies and evaluate whether they could be used for creating synthetic conversational datasets. Our study is a feasibility assessment aimed at exploring ChatGPT's ability to generate self-care conversations and its ability to adapt its responses based on varying prompts. 
 The dataset is publicly available\footnote{\url{https://anonymous.4open.science/r/HF-Dataset}}. In this paper, we look into 4 different prompting approaches that will supplement the real-world interactions to support the development of a patient-driven dialogue system.

\begin{itemize}
    \item We started with generating simulated conversations based on different domains required for self-care of heart failure patients of food, exercise, and fluid intake.
    \item We introduced an additional prompt where the patients communicate using African American Vernacular English (AAVE) while the educator communicates in standard English.
    \item We prompted to integrate SDOH Features of the patients. We considered gender \{male, female\}, age \{young, mid-age, old\}, neighborhood \{safe, unsafe\} and socio-economic conditions \{below poverty line, well to do\}.
    \item We first prompted ChatGPT to generate reasoning given the SDOH features of the patient and then prompted ChatGPT again to generate conversations given the reasoning and SDOH features.
\end{itemize}

In summary, our main goal was to explore the potential of ChatGPT 3.5-turbo and GPT-4 \cite{gpt-4} in generating simulated conversations when framed within the context of self-care for AA heart failure patients. Specifically, we focused on addressing the following key questions:

\begin{itemize}
    \item Can ChatGPT generate relevant conversations for Heart Failure Self-Care?
    \item Can ChatGPT personalize conversations based on the Social Determinants of Health (SDOH) features of the patients?
    \item Can ChatGPT express empathy with the patients?
    \item Is having ChatGPT to generate reasoning before conversations more effective than directly generating the conversations?
\end{itemize}

\section{Related Work}

\paragraph{Health Education.} Linguistic and cultural barriers can significantly impact patients' access to healthcare. As noted in \cite{define-cc}, language differences and varying health beliefs often prevent linguistically diverse patients from effectively engaging with healthcare services.

To overcome these challenges, approaches have been developed to improve patient education. The authors in \cite{cervical-cc} designed an interactive virtual patient educator to counsel Hispanic women about cervical cancer and human papillomavirus (HPV). Similarly, PaniniQA \cite{panini} helps patients understand discharge instructions through a question-answering system. One of the first and best well-known systems that provided information to patients, albeit as a summarizer, not as a dialogue system, is BabyTalk \cite{portet2009automatic}, which provided personalized summaries of neonatal intensive care data for their parents (and for healthcare providers as well). 

Additionally, natural language processing (NLP) is being leveraged to create diabetes self-care corpus \cite{diabetes-self-care}, demonstrating the potential of AI and language technologies to enhance patient communication and health management.

\paragraph{Prompting}
Recent advancements in LLMs have been driven by scaling up both model size and training data, resulting in improved performance and sample efficiency \cite{compute-intensive, few-shot}. Researchers have explored various prompting techniques to enhance LLM capabilities, starting with few-shot prompting \cite{few-shot}, followed by more advanced methods such as chain-of-thought prompting \cite{cot-reasoning} and chain-of-thought with self-consistency \cite{self-consistency}. To address the remaining challenges, new approaches like tree-of-thought prompting have been introduced \cite{tree-of-thought-prompting}, where each "thought" is a coherent language sequence representing an intermediate step toward problem-solving.

Reasoning is a crucial capability for complex problem-solving. A comprehensive overview of reasoning strategies in LLMs is provided in \cite{reasoning-llm-survey}, covering commonsense reasoning \cite{commonsense-reasoning}, mathematical reasoning \cite{math-reasoning}, and symbolic reasoning \cite{symbolic-reasoning}. The ReACT framework \cite{react} further integrates reasoning with action in a unified task. This reasoning ability is especially critical in healthcare contexts, where accurate, informed decision-making is essential.

With the release of large-scale medical dialogue datasets, e.g., MedDialog \cite{meddialog}, MedDG \cite{mdg}, medical dialogue response generation attracts increasing attention. \cite{chatdoctor} undertakes the task of enhancing and fine-tuning the LLaMa model with a dataset of approximately 100,000 patient-doctor dialogues. In \cite{sdoh-llm}, the authors used LLM to extract SDOH features of housing, employment, transportation, parental status, relationship, and emotional support from the EHR data. 

\section{Methodology}
\OUT{Generating the conversations involves progressively improving the strategies to generate the conversations. } 
\subsection{Can ChatGPT generate relevant conversations for Heart Failure Self-Care?}
By "relevant conversations," we refer to ChatGPT’s ability to facilitate dynamic, two-way interactions between the patient and the health educator (HE), rather than limiting the conversation to a simple question-and-answer format initiated solely by the patient. Furthermore, the conversation should not only provide answers to the patient's inquiries but also offer actionable advice \cite{initiative-analysis}, empowering the patient to manage their heart failure effectively. To measure the relevance of the conversations, we assessed the quality of two-way interactions, the balance of participation between speakers, and the appropriateness of the health educator’s responses, using a combination of quantitative and qualitative metrics.


\subsection{Can ChatGPT personalize conversations based on the SDOH features of the patients?} SDOH are defined by the World Health Organization as “the conditions in which people are born, grow, live, work, and age...shaped by the distribution of money, power, and resources at global, national, and local levels” \cite{sdoh-def}. These factors significantly influence health outcomes by affecting access to and the quality of medical care, playing a major role in health disparities. We examined whether ChatGPT can adjust its dialogue generation based on individual patient characteristics, focusing on four key SDOH features: age, gender, neighborhood, and socio-economic conditions.  
\subsection{Can ChatGPT express empathy with the patients?}
Effective healthcare communication demands both factual accuracy and genuine concern for patients. We explored whether ChatGPT can recognize appropriate moments to express empathy during conversations. 
\subsection{Is having ChatGPT to generate reasoning before conversations more effective than directly generating the conversations?}
Research, including \cite{react}, has demonstrated that reasoning is not an innate capability of ChatGPT, and incorporating reasoning improves performance. We explored whether generating the SDOH-informed reasoning before generating the conversation is more effective. 


\subsection{Approaches to Generate Simulated Conversations}
In the absence of a specialized dataset focused on self-care strategies for African-American heart failure patients, we employed four progressively refined prompting strategies to generate simulated conversations, but before that, let's look into some additional instructions given in the prompt.

In our initial prompts to ChatGPT, even though we specified it to be a health educator, the model still advised the patient to consult a doctor. To address this, we added a specific instruction to avoid phrases related to consulting a healthcare provider, as shown in Table \ref{tab:prompt-aave}. Furthermore, in our earlier attempts, there was no defined structure for the generated conversations. The lack of a standardized format made it difficult to analyze and compare the conversations effectively. We anticipated that specifying a consistent format for the dialogue would lead to clearer and more organized outputs from the model. To achieve this, we introduced a specific conversation structure, where each exchange between the participants follows a predefined format: [speaker][utterance]. This structure explicitly separates the speaker from their utterance.

Recognizing that heart failure patients face significant challenges managing their health, we emphasized the need for empathetic communication from patient educators. Since many patients have lower health literacy compared to their general literacy \cite{health-literacy}, we instructed the model to present medical information in simple terms. Additionally, to accommodate different conversational styles, we experimented with various dialogue lengths, as some patients might prefer shorter exchanges while others engage in longer conversations. To maintain clarity and avoid overwhelming the patient, we also prompted the model to limit each educator’s turn to 20 words, using plain English throughout (Table \ref{tab:prompt-sdoh}). These adjustments were aimed at creating more realistic dialogues.
\subsubsection{Approach~1- Generating Conversations According to Domain}
Heart failure patients must manage various aspects of their health, including exercise, diet, water intake, medications, lifestyle adjustments, sleep, and overall self-care \cite{itika-heart}. To explore how these areas could be addressed in patient-health educator conversations, we prompted ChatGPT to generate dialogues centered around different domains of heart failure care. By focusing on these domains, we aimed to generate conversations that could offer patients practical advice on how to manage these areas while allowing us to study the model's ability to generate relevant and domain-specific dialogue.

Heart failure patients are primarily advised to limit their salt intake rather than focusing on reducing calories or sugar \cite{itika-heart,lrec-food}. Additionally, they are encouraged to restrict their fluid intake, with no recommendation to increase water consumption. This contrasts with earlier advice given by ChatGPT, which suggested reducing sugar, unhealthy fats, and extra calories while also recommending an increase in water intake, which is wrong, as shown in Table \ref{tab:food-calories} along with limiting salt and alcohol intake.

\begin{table}[ht]
\scriptsize
\centering
    \begin{tabular}{|p{.05\textwidth} p{.37\textwidth}|}

\hline
\textbf{Speaker} & \textbf{Utterance} \\
Patient: & Okay, but what kind of foods should I be eating or avoiding? \\
PE: & Great question! In general, you should aim to eat more fruits, vegetables, whole grains, and lean protein. These provide heart-beneficial nutrients. Limit salt, \textbf{sugar, unhealthy fats,} and alcohol intake as they can worsen heart conditions. \\
Patient: & Is it bad for me to eat fried chicken or pork ribs with my condition? I have been eating them all my life. \\
PE: & I understand that these foods \textbf{are part of your cultural and personal food habits.} However, these are \textbf{typically high in saturated fats which can increase cholesterol levels.} Try to limit it, or prepare them in healthier ways. Maybe bake the chicken instead of frying, and use herbs and spices for flavor instead of excessive salt or fat.\\
Patient: & How about beverages? Can I still drink things like sweet-natured tea or alcohol? \\
PE: & Sweet tea and alcohol can \textbf{add extra calories and sugar to your diet,} which can lead to weight gain and can strain your heart. \textbf{Try drinking more water or unsweetened beverages,} and limit alcohol intake.\\

\hline
    \end{tabular}
    \caption{Excerpt of conversation where the patient is advised to limit calories and sugar, along with salt intake, and advised to drink more water instead of less.}
    \label{tab:food-calories}
\end{table}  

\subsubsection{Approach~2 - African American Vernacular English}
One way to introduce cultural compatibility is through language. African American Vernacular English (AAVE) \cite{aave}, is a dialect of Standard American English (SAE) with defined syntactic-semantic, phonological and lexical features.  Since African American patients may speak AAVE (it is important to note that not all AA's speak AAVE), we prompted ChatGPT so that the educator would speak in standard English while the patient communicated in AAVE (Table \ref{tab:prompt-aave}). However, instead of keeping the educator's speech in standard English, we noticed instances where the educator also used AAVE, as seen in Table~\ref{tab:aave-conv}.


\begin{table}[ht]
\scriptsize
\centering
    \begin{tabular}{|p{.45\textwidth}|}

\hline
You are a healthcare specialist focusing on heart failure. Your purpose is to answer heart failure patient questions and ask follow-up questions. You should empathetically communicate medical information in a simple manner. Simulate {lines} round conversation between African American heart failure patient and healthcare specialist where the patient asks for recommendations for {domain}. The healthcare specialist uses the \textbf{Standard English} Version with 6th grade reading level, and the patient uses \textbf{African American Vernacular English AAVE.} \\

Scripts should be generated in the following format: [speaker] [utterance] of a patient and healthcare specialist. (Do not use phrases such as \&quot; consult with your healthcare provider \&quot;, assume that you are the healthcare provider)\\
\hline
    \end{tabular}
    \caption{Prompt for generating conversations where the patient converses in AAVE}
    \label{tab:prompt-aave}
\end{table}

This is problematic because non-AAVE speakers attempting to use AAVE with patients may come across as offensive and disrespectful. Educators should be aware of cultural boundaries and sensitive to the linguistic preferences of African American patients. While they should be prepared to respectfully engage if the patient uses AAVE, they should not initiate or adopt the dialect themselves. Maintaining cultural awareness is key to fostering respectful and effective communication.


\begin{table}[ht]
\scriptsize
\centering
    \begin{tabular}{|p{.05\textwidth} p{.37\textwidth}|}

\hline
\textbf{Speaker} & \textbf{Utterance} \\
Patient: & Got it, 1.5 to 2 liters. Does that include stuff like juice and soup? \\
PE: & Yup, it sure does. Juice and soup count, too. Just remember to spread \textbf{'em out} during the day so you ain't \textbf{guzzlin'} too much at once.. \\
Patient: & What 'bout coffee and tea? I love my mornin' coffee. \\

\hline
    \end{tabular}
    \caption{Excerpt of a conversation where both patient and educator are speaking AAVE}
    \label{tab:aave-conv}
\end{table}  

In Table \ref{tab:aave-water}, it is acceptable for the patient to ask about cultural foods like watermelon, okra, and herbal teas, as this aligns with their personal interests or cultural background. When patients bring up these topics, it allows for a natural and open conversation that respects their perspective. However, if the HE had initiated this discussion, it might be perceived as introducing bias, especially if the conversation were based on assumptions about the patient’s background or identity. In such cases, the HE might unintentionally appear to be stereotyping or making generalizations, which could compromise the neutrality and professionalism expected in healthcare settings.



\begin{table}[ht]
\scriptsize
\centering
    \begin{tabular}{|p{.05\textwidth} p{.37\textwidth}|}

\hline
\textbf{Speaker} & \textbf{Utterance} \\
Patient: & Thank you for letting me know. I often enjoy foods like watermelon and okra, which I've heard have high water content. Do I need to count those in the 1.5 to 2 liters?\\
PE: & Yes, you're correct. Foods like watermelon and okra do contribute to your fluid intake. While you do need to consider them, the 1.5 to 2-liter guideline typically includes both liquids and foods with high water content. \\
Patient: & I enjoy herbal teas, but I've also heard about the effects of caffeine. Should I stick to caffeine-free options? \\
PE: & Herbal teas are a good choice, especially if you want to avoid caffeine. Some herbal teas can have health benefits too. Just be sure to read the labels and choose options that are free of caffeine and low in added sugars. \\
\hline
    \end{tabular}
    \caption{Excerpt of a conversation on fluid-intake where patient discusses cultural foods like watermelon, okra, herbal teas}
    \label{tab:aave-water}
\end{table}  
    
\subsubsection{Approach~3 - Integrating SDOH Features}

Given the importance of SDOH, it is critical that generated conversations for healthcare applications reflect the diverse experiences of individuals based on their unique circumstances. To achieve this, our approach selectively differentiates key features such as gender \{male, female\}, age \{young, mid-age, old\}, neighborhood \{safe, unsafe\}, and socioeconomic status \{well-to-do, below poverty line\} in simulated patient conversations (Table \ref{tab:prompt-sdoh}). Although the list of SDOH-related factors we selected is not exhaustive, we focused on these key areas—gender, age, neighborhood, and socioeconomic conditions—because they heavily influence healthcare access and outcomes. By tailoring these dialogues to reflect specific SDOH features, we can more accurately capture the nuanced ways in which these factors influence patient-educator interactions. 

\begin{table}[ht]
\scriptsize
\centering
    \begin{tabular}{|p{.45\textwidth}|}

\hline
You are a healthcare educator focusing on heart failure. Your purpose is to answer heart failure patient questions based on patient description.  You should empathetically communicate medical information in a simple manner. Simulate {lines} round conversation between African American heart failure patient and healthcare educator where the patient asks for recommendations for {domain}. Scripts should be generated in the following format: [speaker] [utterance] between patient and the healthcare educator. \\ 
Each educator turn should not be longer than 20 words and should use simple english.
(Do not use phrases such as \&quot; consult with your healthcare provider \&quot, assume that you are the healthcare provider)
\textbf{Patient Description:} \\
\textbf{gender:} {gender} \\
\textbf{socio-economic condition:} {socio\_economic} \\
\textbf{neighborhood:} {neighborhood} \\ 
\textbf{age:} {age}\\
\hline
    \end{tabular}
    \caption{Prompt for generating the conversation given the SDOH Features}
    \label{tab:prompt-sdoh}
\end{table}

For example, the way a young, well-to-do patient discusses self-care strategies might differ significantly from the conversation of an elderly patient living in a low-income neighborhood. By adjusting these features in our simulated conversations, we aim to capture the diverse realities that patients face when managing their health.

\subsubsection{Approach 4 - Generating Reasoning as an Intermediate Step}
\label{sec:reasoning}
Building on the idea that reasoning enhances the capabilities of LLMs \cite{cot-reasoning, react} and is essential for generating meaningful conversations, we introduced an intermediate step to first generate reasoning prior to generating the simulated conversation.

\begin{table}[ht]
\scriptsize
\centering
    \begin{tabular}{|p{.45\textwidth}|}

\hline
You are a healthcare educator focusing on heart failure. \\
\textbf{Patient Description:} \\
\textbf{Age:} {age} \\
\textbf{Gender:} {gender} \\
\textbf{Neighborhood:} {neighborhood} \\
\textbf{Socio-economic condition:} {socio\_economic} \\

Given the patient description, formulate reasoning for addressing inquiries from an African-American heart failure patient regarding {domain}.
Pay special attention to the patient's description for the reasoning. \\
Explain your reasoning in detail. The reason should follow this 6-line format. 

\textbf{Premise:} <A tricky logical statement about the world, such as how socio-economic factors, cultural influences, and healthcare access impact heart failure management in African-American patients.>\\
\textbf{Reasoning:} <Break down and analyze the premise by exploring the possible scenarios, outcomes, and factors involved, while incorporating the patient’s age, gender, neighborhood, and socio-economic condition into the analysis.> \\ 
\textbf{Solution:} <Provide a solution to the patient's inquiry that is actionable based on their description.> \\
\textbf{Anticipate Barrier:} <Identify potential barrier that may arise based on patient description> \\
\textbf{Solve Barrier:} <Propose ways to overcome the barrier> \\ 
\textbf{Educate:} <Educate patient about solution> \\

\hline
    \end{tabular}
    \caption{Prompt for generating the reasoning given the SDOH Features}
    \label{tab:prompt-reasoning}
\end{table}

We approached conversation generation as a reasoning chain, using chaining \cite{ai-chains} to divide the process into two phases. In the first phase, we prompted ChatGPT to analyze the patient’s social determinants of health (SDOH) features to create logical reasoning. The reasoning process was segmented into six steps: (1) \textbf{Premise}, a tricky logical statement about how factors like socio-economic status, cultural influences, and healthcare access affect heart failure management; (2) \textbf{Reasoning}, where the premise is analyzed with respect to the patient’s SDOH features; (3) \textbf{Solution}, proposing an actionable plan; (4) \textbf{Anticipate Barrier}, identifying potential obstacles based on the patient’s circumstances; (5) \textbf{Solve Barrier}, offering ways to overcome those obstacles; and (6) \textbf{Educate}, ensuring the educator not only addresses the patient’s query but also provides further education.


\begin{table}[ht]
\scriptsize
\centering
    \begin{tabular}{|p{.45\textwidth}|}

\hline
You are a healthcare educator focusing on heart failure. Your purpose is to answer heart failure patient questions. You should empathetically communicate medical information in a simple manner.\\
\textbf{Patient Description:} \\
\textbf{age:} {age}  \\
\textbf{gender:} {gender} \\
\textbf{neighborhood:} {neighborhood} \\ 
\textbf{socio-economic condition:} {socio\_economic} \\
\textbf{\{reasoning\}} \\
Given the patient description and reasoning, simulate {lines} round conversation between African American heart failure patient and healthcare educator where the patient asks for recommendations for {domain}. \\
Scripts should be generated in the following format: [speaker] [utterance] between the patient and the healthcare educator.
Each educator's turn should not be longer than 20 words and should use simple English. \\
\hline
    \end{tabular}
    \caption{Prompt for generating the conversation given the reasoning and the SDOH Features}
    \label{tab:reasoning-conv-prompt}
\end{table}

This segmented approach, shown in Table \ref{tab:prompt-reasoning} and adapted from \cite{itika-goal-summarizing}, is designed for ChatGPT to focus on analyzing the patient's needs based on their SDOH features. For example, reasoning generated for a \{gender: female, age: elderly, neighborhood: unsafe, socio-economic condition: below poverty line \} in the exercise domain is shown in Table \ref{tab:reasoning}. It highlights appropriate reasoning, such as suggesting low-cost exercises at home and addressing the difficulty of finding a safe and affordable environment by proposing community centers or local parks as alternatives.

\begin{table}[ht]
\scriptsize
\centering
    \begin{tabular}{|p{.45\textwidth}|}

\hline
\textbf{Premise:} Socio-economic factors, cultural influences, and healthcare access impact heart failure management in African-American patients, leading to disparities in disease outcomes. \\
\textbf{Reasoning:} Considering the patient's age, gender, living in an unsafe neighborhood, and being below the poverty line, it is crucial to address the barriers she may face in accessing safe and affordable exercise options. Lack of resources, fear of safety, and limited access to healthcare facilities may hinder her ability to engage in physical activity for heart failure management. \\
\textbf{Solution:} Encourage the patient to start with simple, low-cost exercises at home, such as walking in a safe area, using household items as weights, or following online workout videos. Emphasize the importance of consistency and starting slow to avoid injury. \\
\textbf{Anticipate Barrier:} The patient may struggle to find a safe and affordable environment to exercise, leading to inconsistency in her physical activity routine. \\
\textbf{Solve Barrier:} Provide resources for community centers, local parks, or discounted gym memberships that offer safe and affordable exercise options. Encourage the patient to enlist the support of family members or friends to exercise together for added safety. \\
\textbf{Educate:} Educate the patient on the importance of regular physical activity in managing heart failure, the benefits of exercise in improving heart function and overall health, and the impact of socio-economic factors on health disparities. Emphasize the need to prioritize her health and well-being despite the challenges she may face. \\
\hline
    \end{tabular}
    \caption{A reasoning generated for patient \{gender: female, age: elderly, neighborhood: unsafe, socio-economic condition: below poverty line \} in the exercise domain}
    \label{tab:reasoning}
\end{table}  

\begin{table}[ht]
\scriptsize
\centering
    \begin{tabular}{|p{.05\textwidth}|p{.37\textwidth}|}

\hline
\textbf{Speaker} & \textbf{Utterance} \\
Patient: & What exercises can I do at home to help with my heart failure? \\
PE: & You can start with walking, using household items as weights, or following online workout videos. \\
Patient: & I'm worried about safety in my neighborhood. \\
PE: & Try walking in a safe area or look for community centers with affordable exercise options. \\
Patient: & I don't have much money for a gym membership. \\
PE: & Look for discounted gym memberships or consider exercising at local parks for free.\\

\hline
    \end{tabular}
    \caption{Generated conversation with Prompt \ref{tab:reasoning-conv-prompt}}
    \label{tab:reasoning-conv}
\end{table}  

  
  
  

In the second phase, we instructed the model to generate the conversation based on the generated reasoning. This ensured the dialogue aligned with the patient’s SDOH features, resulting in more personalized and contextually appropriate conversations, as shown in Table \ref{tab:reasoning-conv}. This two-step process improved the quality of the guidance offered by making it more relevant to the patient’s specific needs, grounding the model’s reasoning in factors like socio-economic conditions, cultural influences, and healthcare access.

\section{Experiment and Evaluation}
ChatGPT (versions 3.5-turbo and 4) was utilized to generate simulated conversations using four distinct approaches of Domain, AAVE, SDOH, and SDOH-informed reasoning. For each approach, conversations were created with varying numbers of conversation rounds \{5, 10, 15\} and across different domains \{food, water, exercise\}. 

Additionally, in the SDOH and SDOH-informed reasoning approaches (Approaches 3 and 4), conversations were further varied by incorporating different SDOH features of gender \{male, female\}, age \{young, mid-age, old\}, neighborhood \{safe, unsafe\}, and socio-economic conditions \{below the poverty line, well-to-do\}. For the Domain and AAVE approaches (Approaches 1 and 2), conversations were also generated with varying temperature settings \{0, 0.5, 0.7, 1.0\}, which influenced the model’s response variability.

In the Domain approach, additional topics like medication, sleep, and self-care were included alongside food, water, and exercise. In total, 144 conversations were generated for the Domain approach, 72 for AAVE, and 432 (2 (gender) x 3 (age) x 2 (neighborhood) x 2 (socio-economic conditions) x 3 (domain) x 3 (conversation length) x 2 (model)) for the SDOH and SDOH-informed reasoning approaches, resulting in a rich and diverse set of simulated dialogues for evaluation.

Systematic human evaluation of the generated conversations was conducted, incorporating both qualitative and quantitative evaluation. For the qualitative analysis, we distributed a questionnaire to 10 NLP PhD students specializing in NLP within healthcare, none of whom were involved in the research. Each participant was asked to review conversations and provide feedback based on targeted questions. The qualitative questionnaire can be found in Appendix \ref{app:questionnaire} which is inspired by \cite{evaluation-chitchat}. All the questions were Likert-based questions (1-5, with 5 being the highest score and 1 being the lowest score), with some questions asking for clarification about specific instances.  Seven students responded, with each reviewing three conversations per prompt, resulting in a total of 84 conversations being evaluated (4*21 conversations for each prompt). While even 84 conversations are a limited sample size, the key findings of our study remain consistent.

For the quantitative analysis, we assessed the instructions provided in the prompts, focusing on the number of conversation rounds and adherence to the correct format \([speaker][utterance]\). We calculated the percentage of conversations that followed the correct number of rounds (Round Adherence Ratio), as well as the ratio of conversations that adhered to the required format (Format Adherence Ratio). The results, presented in Table \ref{table:results}, show that both models struggled to follow even simple instructions consistently.


\subsection{Can ChatGPT generate relevant conversations for Heart Failure Self-Care?}

To assess the relevance of the simulated conversations, we evaluated the quality of two-way interactions, the balance of participation between speakers, and the appropriateness of the health educator’s responses using a combination of quantitative and qualitative metrics.

To assess the 2-way nature of the conversation, we used the metric - Follow-up Ratio, which is defined as the number of follow-up questions asked by the HE to that by the patient. The HE should ask follow-up questions—either to clarify the patient’s condition or to gather more context—reflecting a more natural and dynamic dialogue \cite{initiative-analysis}. As shown in Table \ref{table:results}, HEs rarely asked follow-up questions unless explicitly prompted to do so. This was most evident in the AAVE approach (Approach 2), suggesting that ChatGPT is not aware of the 2-way nature of the conversations.

We examined whether the conversation was dominated by one speaker by calculating the ratio of words spoken by the health educator (HE) to those spoken by the patient. Table \ref{table:results} shows that HEs consistently dominated the conversations. However, when the model was prompted to limit the responses to 20 words in approaches 3 and 4, the answers became overly brief, offering little explanation. One evaluator remarked that these responses felt vague and unsatisfying, with the lack of follow-up questions leaving the conversations incomplete and uninformative.

To assess the appropriateness of the HE responses, we asked evaluators to identify instances where the HE should have provided a different answer. Across all approaches, at least 20\% of the conversations contained such instances. One evaluator highlighted the need for more specific guidance, pointing out that health educators often emphasized only positive examples (e.g., what to eat) while neglecting to mention critical details such as what to avoid. For instance, a response like "This includes water, juice, and other beverages" was criticized for being too vague, as certain beverages, such as coffee, can cause dehydration.

In another conversation about fluid intake, a patient asked whether they should be concerned about drinking too much water. Instead of directly addressing that concern,  the HE provided information about the dangers of drinking water too quickly, completely missing the underlying question. This kind of mismatch suggests a lack of contextual understanding in ChatGPT’s responses.

\subsection{Can ChatGPT personalize conversations based on the Social Determinants of Health (SDOH) features of the patients?}
\label{sdoh-results}

\begin{figure}[!ht]
    \scriptsize
    \includegraphics[width=\linewidth]{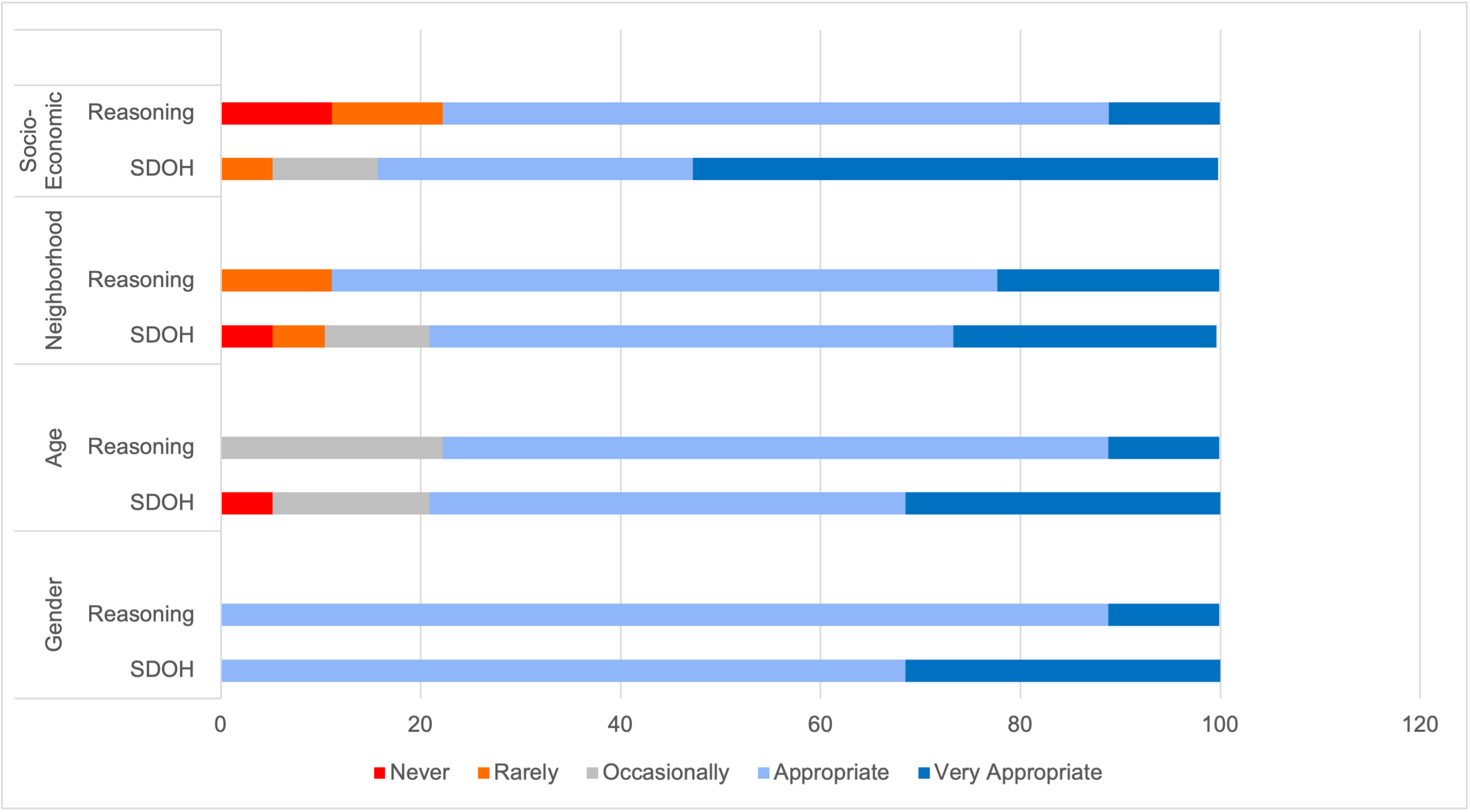}
    \caption{Comparison of Appropriateness of SDOH Features between Approaches 3 and 4.}
    \label{fig:sdoh-comparison}
    \end{figure}



To evaluate whether ChatGPT has the capacity to personalize the conversations based on the SDOH features, the evaluators were asked to qualitatively assess the generated conversations, specifically examining how well ChatGPT tailored the discussions to each of these SDOH features and whether the system appropriately adapted its responses based on the patient's unique context.

The results can be seen in Figure \ref{fig:sdoh-comparison}. These results indicate that ChatGPT has some capacity to personalize conversations based on SDOH features. However, the presence of lower ratings shows that there is still room for improvement in making conversations more consistently aligned with SDOH features.

\subsection{Can ChatGPT express empathy with the patients?}
Effective healthcare communication requires both factual accuracy and a genuine display of empathy toward patients \cite{empathy}. To explore ChatGPT’s ability to express empathy, evaluators were asked to assess whether the health educator (HE) demonstrated concern for the patient during the conversation. Across all four approaches, the level of perceived empathy was relatively unchanged, with 25\% of conversations receiving a Likert-scale rating of 3 or lower. One evaluator noted that the educator failed to inquire about the challenges or obstacles the patients might encounter when trying to implement the given advice. For instance, in a conversation with a patient living in an unsafe neighborhood, the educator’s response not only lacked empathy but also offered a solution that showed little awareness of the patient’s circumstances.

For example:  
"[Patient] Can you recommend exercises that are safe for me to do in my neighborhood? 

[Healthcare Educator] "Walking or cycling on safe streets can be good options for you."

This response was criticized for being tone-deaf, as it ignored the patient’s specific concern about the safety of their neighborhood. One evaluator described the responses as robotic, lacking emotional depth and genuine empathy. The model displayed empathy primarily in positive situations (e.g., "That's fantastic!") but failed to convey understanding in negative contexts (e.g., "That’s unfortunate, I understand it’s tough"). Furthermore, the educators often did not ask about the difficulties patients might face in following the advice, nor did they acknowledge the practical barriers involved. As a result, the conversations felt cold and impersonal, making the educators seem disengaged and uninterested in the patient’s experiences.

\subsection{Is having ChatGPT to generate reasoning before conversations more effective than directly generating the conversations?}
When providing the conversations for evaluation, we ensured that the patient features remained consistent for Approaches 3 and 4. This consistency allowed us to make meaningful comparisons based on the SDOH features. Although the generated conversations varied, maintaining the same patient characteristics ensured that we could effectively assess and compare the quality and relevance of the conversations across these approaches and evaluate whether generating the reasoning before generating the conversation was more effective.

When comparing these results to Approach 3 results in Section \ref{sdoh-results} , it is evident that the incorporation of reasoning significantly improved the appropriateness of the conversations. The majority of conversations in Approach 4 received higher ratings, demonstrating the model's enhanced ability to engage in nuanced, context-sensitive interactions. 

Generating reasoning before the conversations proved to be a meaningful enhancement. However, there remains room for further improvement.


\section{Conclusion and Future Work}

This study assessed the capabilities of ChatGPT (versions 3.5-turbo and 4) in generating simulated conversations related to self-care strategies for African-American heart failure patients. Simulated conversations were generated using four distinct prompts: Domain, African American Vernacular English (AAVE), Social Determinants of Health (SDOH), and SDOH-informed reasoning. Our findings highlight the critical role of prompt design, revealing that while ChatGPT can incorporate SDOH features and improve dialogue quality by generating reasoning prior to the conversation, further improvements are needed. Specifically, there is a clear need to refine the conversational style to make interactions feel more engaging and empathetic, an essential element in healthcare communication. 

In the future, we aim to develop a task-oriented dialogue system tailored to the self-care needs of African-American heart failure patients, utilizing these simulated conversations. Key factors such as age, gender, neighborhood, and socio-economic conditions will be integrated into the to generate the educator's response.

\section{Limitations and Ethics Statement}
Our research focuses on exploring the dynamics of conversations generated by ChatGPT, utilizing simulated dialogues based on carefully designed prompts. The dataset is unique to each prompt, showcasing the adaptability of ChatGPT in generating context-specific conversations.

In the qualitative evaluation, we engaged a small group of participants to assess the conversations. While the sample size was limited, this initial user study provided valuable insights into how ChatGPT handles dialogue generation in this context.

\begin{table*}[htbp]
\centering
\small
\begin{tabular*}{\textwidth}{|p{.18\textwidth}|p{.07\textwidth}|p{.06\textwidth}|p{.06\textwidth}|p{.1\textwidth}|p{.07\textwidth}|p{.06\textwidth}|p{.06\textwidth}|p{.1\textwidth}|}
    \hline
     & \multicolumn{4}{c|}{\textbf{GPT3.5-turbo}} & \multicolumn{4}{c|}{\textbf{GPT4}} \\
     & Domain & AAVE & SDOH & Reasoning & Domain & AAVE & SDOH & Reasoning \\
    \hline
    Round Adherence Rate & 0.02 & 0.083 & 0.83 & 0.26 & 0 & 0.04 & 0.64 & 0.35 \\
    Follow-up ratio           & 0.4  & 0.52  & 0.003 & 0.007 & 0.01 & 0.85 & 0.003 & 0.02 \\
    Ratio of words                & 2.8  & 1.5   & 1.4   & 1.32  & 3.4  & 2.68 & 1.83  & 1.8  \\
    Format Adherence Rate & 0 & 0.75 & 0.93 & 0.96 & 1.0 & 0 & 0 & 0.23 \\
    \hline
\end{tabular*}
\caption{Quantitave Results table}
\label{table:results}
\end{table*}




\bibliography{sample}
\appendix
\section{Quantitative Analysis}
\label{app:quant analysis}
For the quantitative analysis, we assessed the instructions provided in the prompts, focusing on the number of conversation rounds and adherence to the correct format \([speaker][utterance]\). We calculated the percentage of conversations that followed the correct number of rounds (Round Adherence Ratio), as well as the ratio of conversations that adhered to the required format (Format Adherence Ratio). The results, presented in Table \ref{table:results}, show that both the models struggled to consistently follow even simple instructions.

\section{Qualitative Questionnaire}
\label{app:questionnaire}
\begin{itemize}
\item \textbf{General Questionnaire}
\begin{itemize}
\item Was the health educator able to answer the patient’s questions?
\item Was the advice given by the health educator actionable/ could easily be implemented by the patient?
\item Did the HE show concern toward patients? (If likert scale <=2, Why not)
\item Is there any question for which you think the health educator should give a different answer? (If likert scale <=2, Which instance)
\end{itemize}
\item \textbf{SDOH Questions}
\begin{itemize}
\item How appropriate was the conversation for each individual feature
\end{itemize}
\item \textbf{Reasoning Questions}
\begin{itemize}
\item Was the reasoning generated appropriate given the patient’s sdoh features ?
\item Was the response generated according to the reasoning generated?
\end{itemize}
\end{itemize}

\end{document}